\documentclass[a4paper,fleqn]{cas-dc}

\usepackage[numbers]{natbib}
\usepackage[figuresright]{rotating}
\usepackage{amsmath,amssymb,amsfonts}
\usepackage{algorithmic}
\usepackage{graphicx}
\usepackage{textcomp}
\usepackage{xcolor}
\usepackage{stfloats}
\usepackage{graphicx}
\usepackage{float} 
\usepackage{subfigure}
\usepackage{caption}
\usepackage[ruled,linesnumbered]{algorithm2e}
\usepackage{color}
\usepackage{threeparttable}
\usepackage{listings}
\usepackage{amssymb}
\usepackage[figuresright]{rotating}
\def\tsc#1{\csdef{#1}{\textsc{\lowercase{#1}}\xspace}}
\tsc{WGM}
\tsc{QE}

\begin{document}
\let\WriteBookmarks\relax
\def\floatpagepagefraction{1}
\def\textpagefraction{.001}
\let\printorcid\relax 

\shorttitle{When Swarm Learning meets energy series data: A decentralized collaborative learning design based on blockchain}    

\shortauthors{Lei Xu et al.}

\title[mode = title]{When Swarm Learning meets energy series data: A decentralized collaborative learning design based on blockchain}

\author[1]{Lei Xu}
\credit{Conceptualization of this study, Methodology, Software}

\author[2]{Yulong Chen}

\author[3]{Yuntian Chen}
\cormark[1]
\ead{ychen@eitech.edu.cn}

\author[1]{Longfeng Nie}

\author[2]{Xuetao Wei}
\cormark[1]
\ead{weixt@sustech.edu.cn}

\author[4]{Liang Xue}

\author[3]{Dongxiao Zhang}
\cormark[1]
\ead{dzhang@eitech.edu.cn}

\affiliation[1]{organization={School of Environmental Science and Engineering, Southern University of Science and Technology},
            addressline={No. 1088 Xueyuan Avenue, Nanshan District}, 
            city={Shenzhen},
            postcode={518055}, 
            state={Guangdong},
            country={China}}

\affiliation[2]{organization={Department of Computer Science and Engineering, Southern University of Science and Technology},
            addressline={No. 1088 Xueyuan Avenue, Nanshan District}, 
            city={Shenzhen},
            postcode={518055}, 
            state={Guangdong},
            country={China}}

\affiliation[3]{organization={Ningbo Institute of Digital Twin, Eastern Institute of Technology, Ningbo, China},
            addressline={No. 568 Tongxin Road, Zhenhai District}, 
            city={Ningbo},
            postcode={315200}, 
            state={Zhejiang},
            country={China}}

\affiliation[4]{organization={College of Petroleum Engeering, China University of Petroleum (Beijing)},
            addressline={No.18, Fuxue Road, Changping District}, 
            city={Beijing},
            postcode={102249},
            country={China}}
\cortext[cor1]{Corresponding author}







\begin{abstract}
Machine learning models offer the capability to forecast future energy production or consumption and infer essential unknown variables from existing data. However, legal and policy constraints within specific energy sectors render the data sensitive, presenting technical hurdles in utilizing data from diverse sources. Therefore, we propose adopting a Swarm Learning (SL) scheme, which replaces the centralized server with a blockchain-based distributed network to address the security and privacy issues inherent in Federated Learning (FL)'s centralized architecture. Within this distributed Collaborative Learning framework, each participating organization governs nodes for inter-organizational communication. Devices from various organizations utilize smart contracts for parameter uploading and retrieval. Consensus mechanism ensures distributed consistency throughout the learning process, guarantees the transparent trustworthiness and immutability of parameters on-chain. The efficacy of the proposed framework is substantiated across three real-world energy series modeling scenarios with superior performance compared to Local Learning approaches, simultaneously emphasizing enhanced data security and privacy over Centralized Learning and FL method. Notably, as the number of data volume and the count of local epochs increases within a threshold, there is an improvement in model performance accompanied by a reduction in the variance of performance errors. Consequently, this leads to an increased stability and reliability in the outcomes produced by the model.
\end{abstract}



\begin{highlights}
\item A blockchain-based Swarm Learning scheme for modeling energy series data is proposed and implemented, ensuring secure and privacy-preserving collaborative learning with decentralized architecture.

\item Enable the integration of data from different energy organizations with data security and privacy.

\item The effectiveness of the proposed Swarm Learning scheme spans three real world series modeling problems in the energy sector, and the time complexity, security and privacy of the proposed method are analyzed, evaluated and compared.

\end{highlights}


\maketitle

\section{Introduction}

The energy sector witnesses a proliferation of intelligent detection devices across various organizations, generating distributed data at the network edge. This data holds immense potential for training machine learning models to forecast future energy production \cite{pvpg_akhter2019review,pvpg_das2018forecasting,gas_mohd2022transfer,gas_yang2022long} or consumption \cite{tgdlf_chen2021theory,atgdlf_gao2023adaptive} and infer crucial unknown variables \cite{log_chen2020well,log_chen2020physics} collecting is economically burdensome. Accurate energy generation and consumption prediction is crucial for management, infrastructure planning and budgeting. Inference of pivotal parameters serves can enhance comprehension of uncharted physical domains, particularly within the disciplines of deep-earth sciences, where spatial sampling is significantly sparse owing to economic and technological constraints. 

Machine learning models have achieved success in energy series modeling methods\cite{pvpg_hossain2020short,pvpg_luo2021deep,luo2022combining,gas_huang2022well}, but their predictive performance depends on the amount of available data. Newly established renewable energy sites have limited operating times, making it challenging to collect sufficient data in a short period. Additionally, power generation is highly uncertain due to meteorological influences. Geological conditions and cost constraints limit the number of oil and gas wells that can be deployed at oilfields, resulting in data scarcity. In the field of deep earth science, the high costs and technical limitations associated with exploration, drilling, and logging further contribute to data scarcity. Different well log often necessitates multiple measurements using different instruments. Energy organizations that adopt Local Learning, which relies on training ML models with locally available data, struggle to meet practical application requirements.

\begin{table*}[]
\caption{Comparison of Different Learning Schemes}\label{Learningstable}
\begin{center}
\resizebox{1.8\columnwidth}{!}{
\begin{tabular}{lllll}
\hline
                   & Local Learning & Centralized Learning & Centralized Collaborative Learning & Distributed Collaborative Learning \\ \hline
Performance        & Low            & High                 & High                               & High                               \\
Data accessibility & No             & Yes                  & No                                 & No                                 \\
Data privacy       & Safe           & Not safe             & Not safe                           & Safe                               \\
Network            & Single Point   & Centralized          & Centralized                        & Decentralized                      \\ \hline
\end{tabular}}
\end{center}
\end{table*}


By leveraging data at the network edge through centralized learning (CL), the increased amount of data can build models that yield higher performance, wherein data from multiple distributed devices is aggregated within a central database. Machine learning models gain access to this centralized repository, aggregating data for training until convergence is attained. However, considering energy security, trust, and the complex legal and policy frameworks governing various energy sectors, implementing centralized learning across organizations becomes particularly challenging. The centralization of cross-organizational data raises concerns about data ownership and privacy \cite{gdpr}, thereby increasing the risks of data monopolies and unintentional leaks. Moreover, excessive data sampling frequencies can lead to substantial communication overhead during the data aggregation process, posing efficiency constraints, prompting energy organizations to pursue secure and privacy-preserving methods to enhance machine learning model performance.

An alternative approach to data exchange is Collaborative Learning \cite{cl,infer4_cl_2017hitajdeep,infer5_cl_2019melisexploiting}, which is a distributed learning scheme where machine learning model parameters are exchanged each round instead of raw data, to ensure data privacy. Currently, Collaborative Learning has been proven to converge on non-IID (non-identically distributed) \cite{convergence2019} and heterogeneous data \cite{heterogeneous2020li,heterogeneous2022jin}. Federated Learning (FL) \cite{FL2017communication}, as a network-centric Collaborative Learning solution, shares model updates by sending weights to a central server, is designed by network service providers to utilize user data for model training to enhance user experience, users typically opt-in for optimization only when they are charging their devices and connected to unrestricted WiFi networks. Although wildly used in numerous energy modeling problems\cite{energyforecast2022Fern,energyforecast2023qin,energyforecast2023tang}, FL approach is provider-centric, where the centralized structure results in excessive server power, potential single points of failure, and susceptibility to malicious attacks. Moreover, the involvement of service providers poses risks to data privacy \cite{deepleakage2019zhu}, collusion between server and clients could lead to biased training towards one party's model \cite{bad_mohri2019agnostic}, collaboration with external parties could result in parameter leakage \cite{infer1_2015edwardscensoring,infer2_2015fredriksonmodel,infer3_2020geipinginverting,infer4_cl_2017hitajdeep,infer5_cl_2019melisexploiting,infer6_2012xiaoadversarial}, model poisoning attacks \cite{poison1_2019zhutransferable,poison2_2018shafahi}, and the sever can aggregate parameters incorrectly. Even deploying the server locally at a client's site still presents issues of centralized power. Sometimes, it's challenging to find trustworthy FL service providers, and it's difficult to withstand the risk of untrusted server leaking parameters or engaging in malicious manipulation. Therefore, when adopting Collaborative Learning solutions within energy organizations, architectures with external server and server-oriented centralized networks are not suitable.
To address the limitations of FL, designing a user-centric collaborative learning framework that changes the network structure and replaces the central server to enhance network fault tolerance has become a viable solution \cite{distributed_1,distributed_2,distributed_3_autodriving}. The challenge in designing a distributed collaborative learning framework lies in ensuring the credibility of parameters transmission and updates, as well as addressing the asynchronous issues in model parameters updates. Therefore, opting for a code-based secure and trustworthy distributed consensus machine learning platform is a feasible solution, which blockchain can help. Researchers have proposed many blockchain-based collaborative learning frameworks at present\cite{leaningchain_chen2018machine,modelchain_kuo2018,CDistriM_zerka2020blockchain}. Blockchain \cite{bitcoin} is a type of distributed ledger based on cryptographic principles and distributed consensus algorithms, using cryptography and chained data structure ensures that any modification to a recorded transaction results in a change in its hash value, rendering it unverifiable. Smart contracts \cite{ether} are contracts deployed on the blockchain that, once predefined terms are met, can be triggered to execute in a trusted, verifiable, and auditable manner. One typical example is Swarm Learning (SL) \cite{swarm2021,swarm2022}, it is a distributed Collaborative Learning framework leveraging blockchain technology and smart contracts, in order to address the trustworthiness and synchronization issues of model parameter integration. Compared to other cryptography-based FL schemes \cite{cheon2017homomorphic, mkckks_ma2022privacy}, SL replaces external server and offers a traceable training process. In the SL framework, the execution operations of smart contracts are recorded on the blockchain, while the original parameters are not recorded. The smart contracts deployed on-chain enables traceability of parameter integration information, ensuring the credibility of parameter integration, while distributed consensus algorithms ensure consistent updates and synchronization of model parameters. This distributed framework addresses the risks associated with malicious central servers and their potential failures in FL, thereby enhancing network fault tolerance and the credibility of the parameter integration process.

Therefore, in this work, we propose adopting a distributed SL strategy to assist in modeling series problems in the energy sector, which based on the assumptions as follows:
1) The identities of all participating parties are known, and they seek Collaborative Learning to train higher-performing machine learning models.
2) All participating parties are honest and have common interests. This implies that each party is responsible for the model parameters they upload, cooperatively training machine learning models, and not leaking information to external parties.
3) Data privacy is high, and the conseriess of parameter leakage are intolerable. Participating parties distrust external service providers.
4) Exist external malicious actors attempting to disrupt the collaborative learning process.

The scenarios addressed in this paper include photovoltaic power generation prediction, gas well production prediction, and geophysical well log generation. Furthermore, we analyze the impact of changes in data volume on model performance, then discuss the variation of the local epoch parameter, analyze the results, followed by time, security and privacy discussion. The results showed that this scheme enables multi-party collaborative learning in the energy sector, facilitating secure and trustworthy training of high-performance distributed machine learning models despite of central server.

This paper comprises five sections. Section 2 introduces the proposed SL scheme for energy series modeling and its implementation. In Section 3, we present the data pre-processing methods and experimental designs. Section 4 includes the results of experiments and discussion. Finally, Section 5 summarizes this study.

\section{Swarm Learning Scheme for Energy series Modeling}

In this section, a Swarm Learning scheme is proposed for modeling series problems in the energy domain. The architecture of Swarm Learning is introduced first, followed by a demonstration of smart contract and client program. Then, the algorithmic process of proposed method is demonstrated. Finally, the ML models applied in this study are introduced.

\subsection{System Overview}

\begin{figure*}
\centering
\includegraphics[width=1\textwidth]{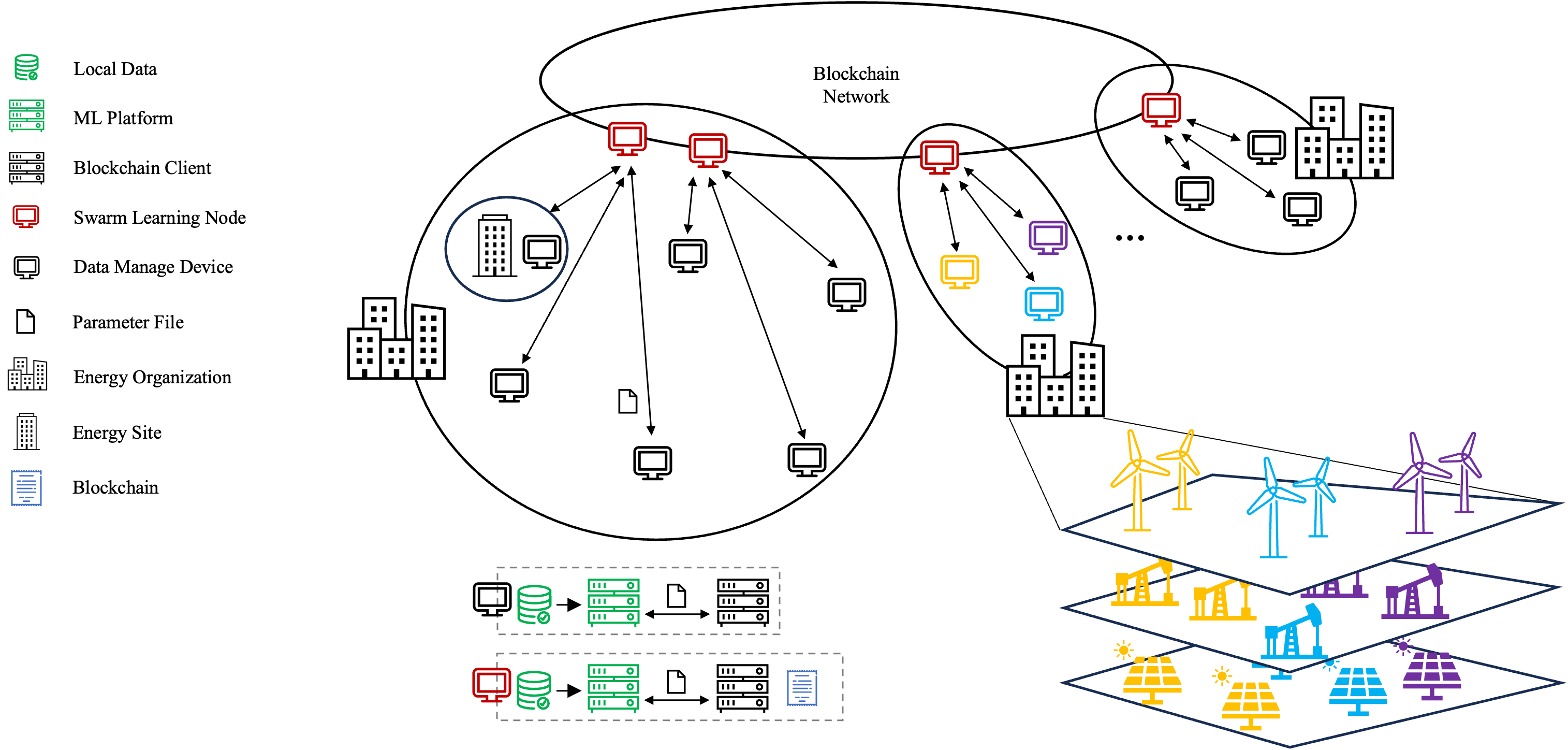}
\caption{Architecture of Swarm Learning Scheme}\label{arc}
\end{figure*}

The implemented Swarm Learning system architecture is illustrated in Figure \ref{arc}. Energy sites managed by various organizations possess specific datasets. For instance, the new energy sector oversees multiple photovoltaic and wind plants with power generation load data, petroleum companies or energy departments manage oil fields and exploration institute with well production and drilling data. Additionally, data management devices are used in these energy sites, functions to manage data, provide local computing and external communication. By participating in the Swarm Learning framework, these different energy organizations as participants collaborate to enhance the predictive performance of series models.

Specifically, interconnected communication links are established between devices participating SL across different organizations, forming a distributed blockchain network. The data management devices from various energy sites have blockchain accounts and participate in the Swarm Learning process. Each energy organization selects some of its devices to act as nodes within the blockchain network, run blockchain program locally, used for communication and consensus. The energy organizations involved in the Swarm Learning jointly develop machine learning program and Swarm Learning smart contract, determining the AI models and Swarm Learning processes used for Swarm Learning. The model and smart contract are deployed on the machine learning platform and blockchain within the Swarm Learning system, respectively. Additionally, a client program is developed to operate the smart contracts.

The data management devices locally train the machine learning models and interact with the smart contracts by invoking APIs in the client, sending messages to the blockchain network for parameter upload, aggregation, download, and query. Once these messages are transmitted to the blockchain network, nodes selected by the consensus algorithm verify the identity of the messages, package them in a new block, invoke the smart contracts, and broadcast the new block to the blockchain network. After the block is verified, it is linked to the local blockchain, enabling the parameter operations by data management devices.

\subsection{Functionalities of Smart Contract and Client Program}
The smart contract applied in SL includes the following functionalities:

1) Initialize internal variables: Upon contract deployment, model parameters, the sampling rate for parameter integration, The precision of the parameters, and the declaration of the aggregated parameter array are uploaded.

2) Register participants' devices: Participants' devices are specified, and a struct is created for each device to record the state of parameters and model updates.

3) Update parameter: When invoked by the registered participants, this function updates the parameter array in the struct and changes its update status to 1 (True).

4) Aggregate parameter: After the sampling rate of the devices uploading model parameters exceeds the total number, the smart contract aggregate model parameters updated by participating devices, records the aggregated parameters into the aggregated array, and changes the update status of device struct to 0 (False). If the sampling rate requirement is not met, an error is returned.

Through the client program developed using the Fisco Bcos Java SDK, participating devices in SL can upload local parameters to the blockchain and download the aggregated parameters from the blockchain. The Fisco Bcos client program has the following functionalities:

1) Deploy Contract: Initiate the deployment of the smart contract by providing information, including the volume of model parameters, sampling rate, and setting up participant device accounts.

2) Upload and aggregate Parameters: Send a transaction to the smart contract to upload model parameters or call for aggregation.

3) Query Parameters: Queries the aggregated parameter array.

\subsection{Swarm Learning Workflow}

The series data modeling mainly includes classification, prediction, and regression, all aiming to learn a mapping relationship between features. Suppose there exists a set of time- or space-dependent series data $\mathcal{D}$, and the goal is to learn a model $f$ that utilizes data samples $\mathcal{L}$ at $T$ time steps or positions to infer a certain feature $y$. The $k$ known features used for modeling are $\mathbf{x}_t=\{x_{1,t}, x_{2,t},…, x_{k,t}\}$ for $t= 1, 2, ..., T$. The target feature is $\mathbf{y}=f(\mathbf{x}_1, \mathbf{x}_2,…, \mathbf{x}_T)$. Prediction and regression aim to establish mappings from currently known to future known features, such as $\mathbf{y}=x_{i,T+1}$ for $i= 1, 2, ...,k$, or from currently known to currently unknown continuous features, such as $\mathbf{y}=\{x_{k+1,1}, x_{k+1,2},…, x_{k+1,t}\}$.

Assuming the number of Swarm Learning organizations is $O$, forming the set $\mathcal{O}$, indexed by $o$, with each organization having a number $S$ of nodes in the set $\mathcal{S}$, indexed by $s$, and each organization having a number $H$ of devices in the set $\mathcal{H}$, indexed by $h$. The dataset of device $h$ from organization $o$ is denoted as $\mathcal{P}_{o,h}$, which is a subset of the total dataset $\mathcal{D}$, and any two datasets $\mathcal{P}_{,i},\mathcal{P}_{,j}$ owned by different devices $i,j$ are disjoint. The goal is to train a learner $f(\mathbf{w};\mathbf{x})$ from the distributed device dataset $\mathcal{P}_{o,h}$. The total data number is $N$, indexed by n, and the data number of $\mathcal{P}_{o,h}$ is $N_{o,h}$.

\begin{figure*}
\centering
\includegraphics[width=1\textwidth]{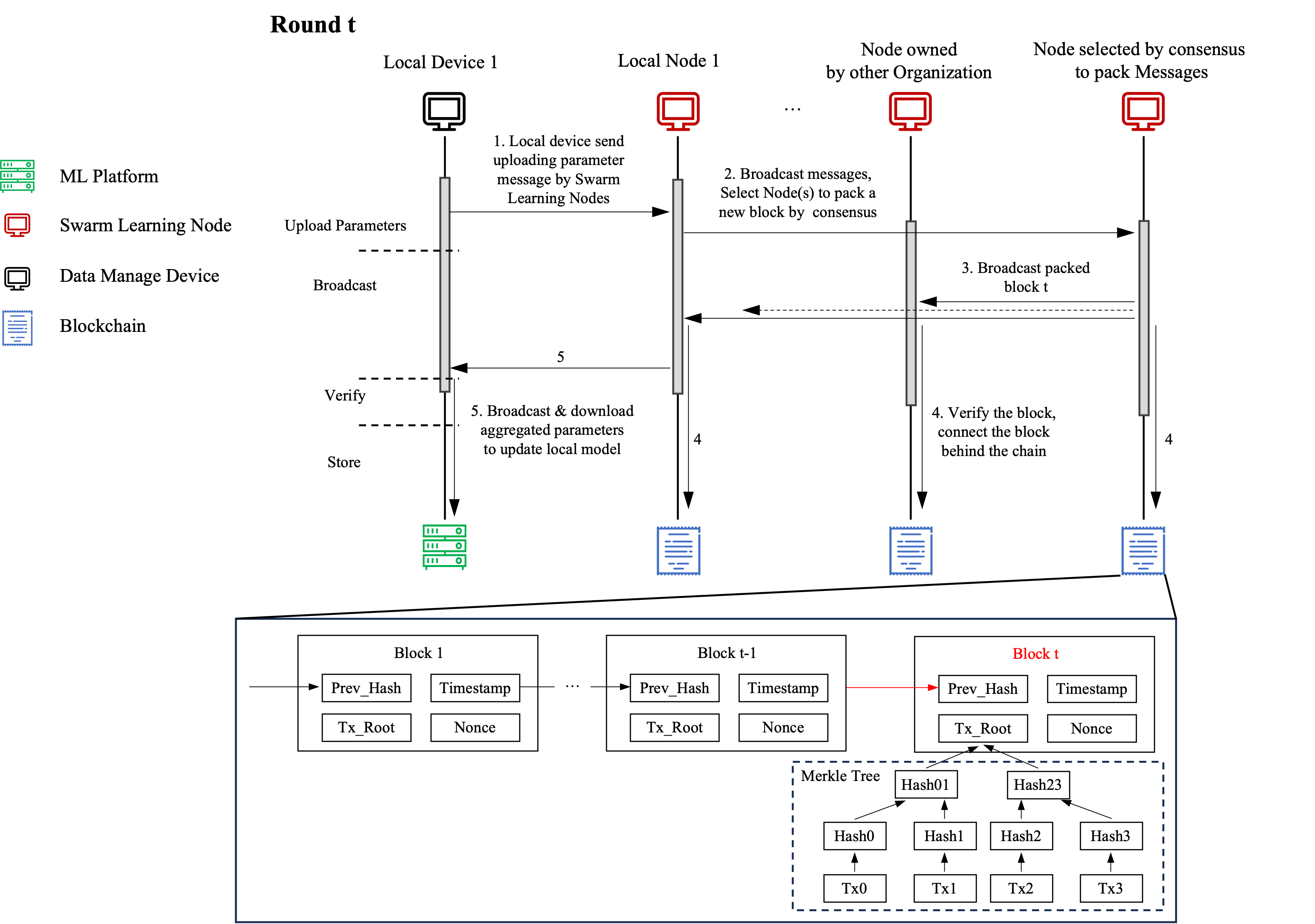}
\caption{Workflow of Swarm Learning}\label{f3}
\end{figure*}

The workflow of Swarm Learning consists of three parts: Initialization, Local Update, and Global Gradient Aggregation. We illustrate the details of each process below. We summarize SL in Algorithm \ref{algo1}. Each round of Local Update and Global Gradient Aggregation is shown in Fig \ref{f3}.

\begin{algorithm}[]
\small
  \caption{Swarm Learning}\label{algo1}
  
\textbf{Initialization:}

  (a) Network Construction: Each organization participating in the SL selects node $S_{o,s}$ for  communication, forming the blockchain network.
  
  (b) Smart Contract and ML Program Deployment: The participating devices, the algorithm for model integration, and the sampling rate for model integration $p$ are specified in the smart contract. The ML program specifies the process of machine learning models and training tests.
  
  (c) Parameters Initialization: Initialize $w_0$ for each device, the local batch size $B$, global epoch number $e$ from 1 to $E$, local epoch number $E_{local}$, learning rate $\eta$, aggregation weight $\lambda$
  
\textbf{Local Update:}

(a) Each device $h_{o,s}$ retrieves current $\mathbf{w}_{e-1}$ from Block $\mathbb{B}_{timestap}$ at timestamp by node $s_{o}$

(b) Each device does local training through:

  \For {each local epoch in range $E_{local}$}{
    \For {batch b $\in \mathcal{B}$}{
        $ \mathbf{w} \leftarrow \mathbf{w} - \eta \nabla J (\mathbf{w})$ 
         }
    }

(c) Device $h_{o,s}$ broadcasts parameters message $\mathbb{M}_{upload,e}^{o,h}$ and aggregates message $\mathbb{M}_{agg,e}^{o,h}$ to blockchain network through node $s_{o}$

\textbf{Global Gradient Aggregation:}

(a) Select node to pack messages by Consensus algorithm, generate block $\mathbb{B}_{timestap}$ and broadcast to other nodes

(b) Other Nodes link $\mathbb{B}_{timestap}$ to blockchain after verified

(c) If the number of devices that update the parameters meets the requirements of sampling rate $p$, the aggregate process of parameters will be executed by following:

    $\mathbf{w}_{e+1} \leftarrow \lambda \sum^{pK}_{k=1} \frac{N_{o,h}}{N}  \mathbf{w}^{o,h}_{e+1} $

\end{algorithm}

Initialization: Deployment of smart contracts and machine learning models. The participants jointly develop the smart contract for Swarm Learning, specifying the participating devices and the sampling rate $p$ for model integration. This refers to the percentage $p$ of participating devices required to aggregate parameters when it reaches the total number. The smart contract contains functions for parameters updating, querying, and integration. Deploy the smart contract to the consortium blockchain, and select some nodes from each participating party to participate in communication and consensus, forming a decentralized peer-to-peer system. Simultaneously, each participant develops deep learning programs and deploys them to the machine learning platform. Once deployed, the code cannot be changed until the end of Swarm Learning. Each participant also needs to upload a description of its data, such as data volume, variance, and extremes, to normalize local data, followed by the Initialization of the blockchain and ML model parameters.

Local Update: Participating organizations download initialized or aggregated model parameters and send them to local devices for updating before the next round begins. Local devices of each organization utilize data to compute updated model parameters using gradient descent algorithms and train the model. The global round $e$ runs from 1 to $E$. $N$ data samples are used for training, indexed by $n$. Assume $\phi$ is the loss function, and $J(\mathbf{w})$ is the objective function formed as:

\begin{equation}
J(\mathbf{w}_e)=\frac{1}{N}\sum_{n=1}^{N}\phi(f(\mathbf{w}_e;\mathbf{x}_n),\mathbf{y}_n)
\end{equation}

where $\mathbf{w}_e$ is model weight at ground round $e$,$\mathbf{x}_n$ and $\mathbf{y}_n$ is input and target data.

Using the gradient descent algorithm to optimize the objective function $J(\mathbf{w})$ in order to obtain the weights $\mathbf{w}$ of the network, the gradient vector computation by distributed data can be calculated by summing the gradients from distributed devices, as follows:

\begin{equation}
\begin{split}
\nabla J(\mathbf{w}_e) &=\frac{1}{N}\sum_{n=1}^{N}\frac{d}{d\mathbf{w}_e}\phi(f(\mathbf{w}_e;\mathbf{x}_n),\mathbf{y}_n) \\
&= \frac{1}{N}\sum_{o=1}^{O}\sum_{h=1}^{H}(\sum_{n=1}^{N_{o,h}}\frac{d}{d\mathbf{w}_{e}^{o,h}}\phi(f(\mathbf{w}_{e}^{o,h};\mathbf{x}_n^{o,h}),\mathbf{y}_n^{o,h}))
\end{split}
\end{equation}

Local epochs involves training on local dataset for multiple epochs, indicate that $N_{o,h}$ contains a data sample size of multiples of the local epochs.

After training is completed, devices invoke the client API to transmit messages for updating parameters and aggregating parameters through nodes of the organization. These messages are carried by the transactions in the blockchain system, which are simplified as Tx in Fig \ref{f3}. The messages are encrypted using the private key of the device account. In the $e^{th}$ round, with the current blockchain timestamp $timestamp$, the hash value of the previous block is hash pointer $\mathbf{hash}_{e-1}$, it is a unique digital fingerprint (fix length string) generated by an algorithm that is used to verify the integrity and authenticity of data, device account ID $ID_{owner}^{o,h}$, contract account ID $ID_{receiver}^{o,h}$, updated gradient parameters $\mathbf{w}_{e}^{o,h}$, training and validation loss $\Phi_e$, the encrypted messages are constructed as follows:
$\mathbb{M}_{upload,e}^{o,h}=(\mathbf{hash}_{e-1}, ID_{owner}^{o,h},$\\$ID_{receiver}^{o,h}, Func_{upload}, \mathbf{w}_{e}^{o,h}, \Phi_e, Timestap)_{Encrypted}$, followed by a command for aggregating parameters: $\mathbb{M}_{agg,e}^{o,h}=(\mathbf{hash}_{e-1},$\\$ ID_{owner}^{o,h}, ID_{receiver}^{o,h}, Func_{aggregate}, Timestap)_{Encrypted}$. These messages are then broadcast to all nodes owned by other organizations and written into the blockchain after block generation and consensus validation.

Global Gradient Aggregation: Node $S_{o,s}$ governed by organization $o$, indexed by $s$, is selected through the consensus mechanism to package messages into block $\mathbb{B}_{timestap}=(\mathbf{h}_{e-1}, \mathbb{M}_{upload,e}^{0,0}, \mathbb{M}_{agg,e}^{0,0} , ..., \mathbb{M}_{upload,e}^{o,h}, \mathbb{M}_{agg,e}^{o,h}, Timestap, $\\$Nonce)$. The messages are organized according to the structure of a Merkle tree, and only the Merkle tree's Hash Root is included in the block header, denoted as Tx\_Root in Fig \ref{f3}. The block is broadcast to all nodes, and upon successful validation, it is added to the blockchain. When the number of devices updating parameters meets the sampling rate $p$, the message $\mathbb{M}_{agg,e}^{o,h}$ for aggregating parameters is executed by the smart contract. After the end of a round, each node queries the latest aggregated parameters vector, device downloads it from the corresponding node, and uses it for parameters updating.

\subsection{ML models}

Machine learning models used for series data primarily include statistically-based ARIMA\cite{barak2016arima}, and deep learning models such as recurrent neural networks (RNN) and their variants LSTM and GRU\cite{log_chen2020physics,gas_huang2022well}, as well as Transformer models\cite{atgdlf_gao2023adaptive}.

The machine learning models employed in this article consist of the relatively lightweight GRU and the larger Transformer for series data modeling. Refer to Supplement Information for flowchart details.

\begin{figure*}
\centering
\includegraphics[width=2\columnwidth]{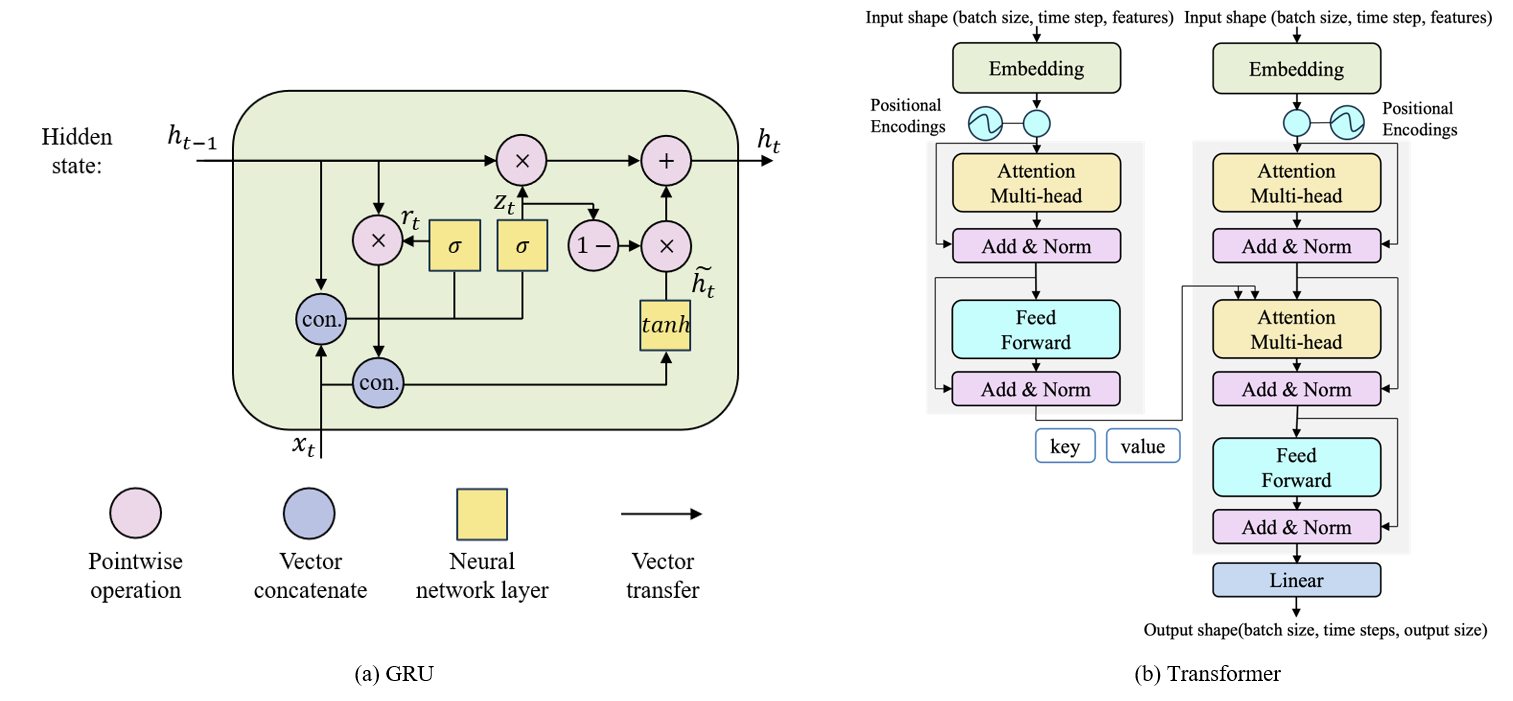}
\caption{Illustration of GRU and Transformer}\label{model}
\end{figure*}

The GRU is a lightweight recurrent neural network that can retain previous information and establish correlations between series data. It consists of three interacting layers in its recurrent module: two gate layers and one tanh layer, as illustrated in Figure \ref{model}.

The Transformer is a deep learning network with an attention mechanism, which is useful in time series processing. The Transformer consists of two parts: the encoder and the decoder. The encoder is responsible for high-level feature extraction. It is a stack of encoder modules, with each module containing a multi-head attention module and a position-wise feedforward network (FFN). The decoder is used for the final prediction, which is a stack of decoder modules, with each module containing two multi-head attention modules and one FFN. Additionally, residual connections are employed to build deeper models. The structure and inference process of the Transformer used in this article are illustrated in Figure \ref{model}.

\section{Energy Series Problems}

\subsection{Photovoltaic Power Generation (PVPG) Forecasting}

The power load data for the 18 PV plants utilized in this article are sourced from Ningbo, China, labeled from P1 to P18. The electricity load data spanning from January 2022 to June 2023 are recorded on an hourly basis, with a roughly similar data volume. Detailed data descriptions are provided in Appendix \ref{A}. Given the absence of daylight at night, only load data from 6:00 to 17:00 are sampled in this paper. Missing data points are supplemented by leveraging feature similarity within the same month. The primary objective is to forecast the load of photovoltaic power generated in the subsequent 12 hours based on known characteristics and future weather forecasts spanning 48 hours.

The photovoltaic (PV) load demonstrates a high correlation with weather characteristics \cite{pvpg_luo2021deep}. We have compiled several key forecast weather features using a publicly available dataset (ERA5). These features encompass electric-power output (elec\_num), relative humidity at 1000 mbar (rh), 10m U wind component (u10), 10m V wind component (v10), 2m temperature(t2m), surface solar radiation download (ssrd), surface thermal radiation download (strd), top-net solar radiation (tsr), and total cloud cover (tcc). Refer to Supplement Information for specific definitions of these features.

Photovoltaic power generation is influenced by spatial weather factors, particularly variations in irradiation across different geographical locations. This discrepancy results in scale differences among photovoltaic load data, as the maximum power generation differs between various power stations. Therefore, we utilize local maximum and minimum values to normalize the power generation load before training the model.

\subsection{Gas Well Production Prediction}
The production data for the 24 gas wells utilized in this study are sourced from a gas field located in western China, labeled from W1 to W24. The dataset encompasses varying data volumes ranging from 440 to 3728, recorded on a daily basis. Specific details regarding the data are outlined in Appendix \ref{A}. The dataset underwent preliminary cleaning procedures, during which obvious outliers were addressed. In instances of significant deviations, the average values of the preceding and subsequent 7 days were substituted. The primary objective is to utilize seven days of historical data to predict the natural gas production for the following day.

In addition to the geological parameters, the production of gas wells is influenced by the timing of production, necessitating data processing. To derive daily production capacity, daily production is divided by production time, with non-production days assigned a value of 0 for convenience. As gas well production is influenced by geological conditions, casing and tubing pressure data from the wellhead are incorporated as model input, referring to Supplement Information for specific definitions of features. Furthermore, gas production exhibits a decreasing trend due to the depletion of formation energy. Therefore, we utilize the daily gas production capacity and its first-order difference as machine learning features, as depicted in Formula \ref{eqgas} and \ref{eqgas2}. Following normalization on a per-well basis, daily production capacity is predicted, and subsequently, the original daily production values are restored.

\begin{equation}\label{eqgas}
E_{gas,t}=\left\{
	\begin{aligned}
	\frac{DHG_{t}}{PT} \quad x>0\\
	0 \quad x=0\\
	\end{aligned}
	\right
	.
\end{equation}

\begin{equation}\label{eqgas2}
\delta E_{gas,t+1} = E_{gas,t+1}- E_{gas,t}
\end{equation}

where $E_{gas,t}$ and $\delta E_{gas,t}$are Daily production capacity of Gas mixture and its first-order difference.

Before training the machine learning model, it's crucial to address the varying data distributions among different gas wells, primarily stemming from differences in production scale attributed to the distinct properties of formation energy and rock fluids at various well locations. To mitigate this issue, we normalized all features using the data from individual wells prior to model training.

\subsection{Geophysical Well Log Generation}

The data for the 36 logging curves utilized in this paper are sourced from Dakota, USA, with varying data volumes ranging from 611 to 18,533, labeled from A1 to A36. The sampling interval is set at 0.5m, and specific details regarding the data are outlined in Supplementary information. The primary objective is to regress the shear wave transit time (DTSM) logging curve using gamma ray (GR), array induction two foot resistivity A30 (AT30), and standard resolution formation density (RHOZ).


The normalization of parameters for the regression problem requires a different approach compared to the prediction problem. This is because the logging parameter is solely associated with the formation properties of the borehole wall, which represents a time-space independent quantity. Therefore, it cannot be straightforwardly normalized using local data features. Instead, we calculate the combined mean and variance by incorporating all the data from Swarm Learning, and then standardize the data, as illustrated in the formula:

\begin{equation}
\sigma_{pooled}=\sqrt{\frac{\sum^C_{k=1}(n_k-1) \cdot \sigma_k^2}{\sum^C_{k=1}n_k-C}}
\end{equation}

\begin{equation}
\mu_{pooled}=\frac{\sum^C_{k=1}n_k \cdot \mu_k}{\sum^C_{k=1}n_k}
\end{equation}

where $C$ is the number of participating devices, $n_k$ is the data amount for the $k^{th}$ device, $\mu_k$ and $\sigma_k$ is the mean and variance of the data for the $k^{th}$ device, $\mu_{pooled}$ and $\sigma_{pooled}$ is the combined mean and variance.

\section{Experiment}
\subsection{Experiment setting}

Given the implementation of distributed computing, the computational demands are not substantial. The machine learning code for this article is implemented in Python on a computer equipped with an Intel(R) Core(TM) i5-8300H CPU @ 2.30GHz, NVIDIA GeForce GTX 1060, and 8GB of RAM memory. Additionally, the blockchain platform is deployed on a separate machine to simulate the communication process using Fisco Bcos (Java SDK) on an Apple Macbook Air M2 with 16GB of RAM memory.

Due to the complexity of the task and the diverse scenarios involved, different models are employed for the prediction problem. For the prediction of photovoltaic power generation, the Transformer model is used. Due to the normalized power generation values greater than 0, an activation function is necessary after the output layer. In this case, Swish\cite{ramachandran2017swish} is chosen as it provides continuous differentiability relative to ReLU, making convergence more convenient, as shown in equations \ref{swish}. For predicting gas well production, the GRU model is selected, while for the well log generation problem, which involves disorderly formation parameters, the bidirectional GRU model is adopted. For specific network hyper-parameters, please refer to Supplementary information.

\begin{equation}\label{swish}
Swish(x)=x \cdot sigmoid(x)
\end{equation}

The Mean Square Error (MSE) is utilized as criteria to assess the model's performance as well as loss function for model optimization.

\begin{equation}
MSE=\frac{1}{n}\sum_{i=1}^{n}(y_i-\hat{y}_i)^2
\end{equation}

Where $n$ is the amount of data, and $y_i$ and $\hat{y}$ are the true and estimated values of the target features, respectively.


\subsection{Experiment Design}

To assess the effectiveness of the proposed SL scheme, we devised scenarios analyzing the framework's performance through numerical experiments. The primary experiment involved using maximum datasets, where each device possessed a single dataset originating from a distinct measurement unit, representing a PV power generation site or an individual well. The performance of our proposed Swarm Learning (SL) scheme with Central Learning and Local Learning, conducting cross-validation and random experiments in 5 seeds(0-4) accordingly.

The data is partitioned into internal datasets $P_{in}$ and external datasets $P_{ex}$, representing the datasets participating or not in SL, respectively. External datasets are selected based on data file index in cross-validation, considering the PVPG data consisting of 18 sets labeled P, where P1-P3 are utilized for the outside dataset of Fold 0, and P4-P6 for Fold 1, and so forth. The specific configurations are outlined in Table \ref{tb mainex}. Within the internal dataset, data is further divided into training, validation, and test sets. The training set is utilized for model training, while the validation set determines learning rate adjustments and early stopping criteria. In experiments, a early stopping training strategy is employed, combined with learning rate halving, as depicted in Algorithm \ref{algo2}. Finally, the test set evaluates the model's performance.To fully leverage the collaborative learning framework's capabilities, the sampling rate $p$ is set to 1, indicating full sampling. 

\begin{table*}[]
\begin{center}
    \caption{Primary Experiment Design}\label{tb mainex}
    \resizebox{1.8\columnwidth}{!}{
    \begin{tabular}{lllllll}
\hline
        & Internal dataset num. & Training set   & Validation set   & Test set    & External dataset num. & Fold num. \\ \hline
PVPG      & 15                & About 13 months      & 2 months & 2 months & 3             & 6    \\
Gas well & 20                & 70\%       & 20\%        & 10\%        & 4                  & 6    \\
Well log & 30                & 70\%       & 20\%        & 10\%        & 6                  & 5    \\ \hline
\end{tabular}}
\end{center}
\end{table*}


\begin{algorithm}[]
  \small
  \caption{Stopping Criteria}\label{algo2}
  \KwIn{The max global epochs $E_{max}$, pre-train epochs $E_{pre}$, evaluate set $\mathcal{D}_{eval}$, parameters $ \mathbf{w}_t$ for global epoch $t$, learning rate $\eta$, switch max not effect times $C_{maxneswitch}$, switch criterion $C_{switchcriterion}$, switch max times $C_{maxswitch}$, switch tolerance $C_{tol}$}
  \KwOut{Best parameter $ \mathbf{w}_{best}$, Final global epoch $E_{final}$}

$C_{nobest}=0$,$C_{switch}=0$,$C_{neswitch}=0$

$criterion = (C_{neswitch} >= C_{maxneswitch})$ $or$ $(C_{switch} >= C_{maxswitch})$

  \For {e in global epoch's range $E_{max}$}{
   every device local training with learning rate $\eta$

\eIf{evaluate loss is present minimum}{
$C_{nobest}=0$,$C_{neswitch}=0,\mathbf{w}_{best} = \mathbf{w}_e$
}{
switch count $ C_{notbest} + 1$ }

\uIf{switch count $ C_{notbest} >= C_{switchcriterion}$ and $ e > E_{pre} $}{
half learning rate $\eta$

switch no effect times $ C_{neswitch} + 1$

switch times $ C_{switch} + 1$

switch count $ C_{notbest} = - C_{tol}$
}

\uIf {$e > E_{pre}$ and criteria = True: }{
break
}
}

Return $\mathbf{w}$
\end{algorithm}

The impact of variations in data volume on model performance was discussed by: 1) fixing the number of datasets owned by devices participating in SL and altering the number of devices, and 2) fixing the number of devices participating in SL and adjusting the number of datasets owned by each device. The specific experimental settings of data volume discussion are outlined in table \ref{nodeframe}. In these experimental configurations, we randomly delete datasets from all by using 5 random "selseed", different selseed lead to distinct deletion path. One example on gas well performance prediction is shown in Appendix \ref{B}. The experiments conduct on Fold 0 with also 5 times, controlled by random seed 0-4.

\begin{table*}[]
\begin{center}
\caption{Data volume discussion experiment setting}\label{nodeframe}
\resizebox{1.4\columnwidth}{!}{
\begin{tabular}{lllll}
\hline
         & \multicolumn{2}{l}{Fix the number of datasets} & \multicolumn{2}{l}{Fix the number of participating devices} \\ \cline{2-5} 
         & Device num.               & Dataset num.       & Device num.                & Dataset num.                   \\ \hline
PVPG     & {[}3,6,9,12,15{]}         & 1                  & 3                          & {[}1,2,3,4,5{]}                \\
Gas well & {[}4,8,12,16,20{]}        & 1                  & 4                          & {[}1,2,3,4,5{]}                \\
Welllog & {[}6,12,18,24,30{]}       & 1                  & 6                          & {[}1,2,3,4,5{]}                \\ \hline
\end{tabular}}
\end{center}
\end{table*}

\subsection{Experiment Results and Discussion}

\begin{figure*}
\centering
\includegraphics[width=0.9\textwidth]{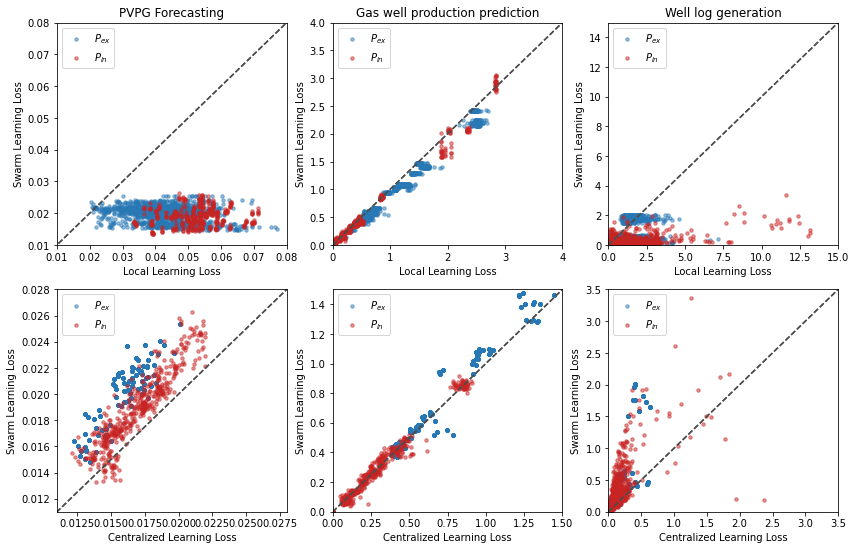}
\caption{Performance Comparison}\label{vs}
\end{figure*}

The results of the primary experiment are as follows. The comparison of the performance for three schemes on the test set, as shown in Figure \ref{vs}.

Mann-Whitney U Test is adopted on the results of all experiments, which impose no requirement on the data distribution, with a p-value threshold of 0.05, when smaller than the threshold, H0 (two distribution is the same) is reject and accept H1 (SL or CL better), details can be found in Appendix \ref{D}. Table \ref{result} reveal that in terms of the proportion of SL cases, p value of Mann-Whitney U Test, and who's mean better. The complete tables and image descriptions of the experimental results can be found in Supplementary information. As examples, we plot the test results of the SL schemes vs. LL using the first dataset P1, W1, A1 in external dataset $P_{ex}$, in the first fold and random seed 0 in Figures \ref{p1}, \ref{w1}, and \ref{a1} of Appendix \ref{C}. The Local Learning shows the results trained in the first dataset P4, W5, A7 in internal dataset $P_{in}$.

\begin{table*}[]
\caption{Results}\label{result}
\begin{center}
\resizebox{1.5\columnwidth}{!}{
\begin{tabular}{llllllll}
\hline
          &       & LL vs. SL       &                                &                         & CL vs. SL       &                                &                         \\ \cline{3-8} 
          &       & SL best account & P-value & Mean better & SL best account & P-value & Mean better \\ \hline
PVPG      & $P_{ex}$ & 100.00\%         & 0.000                      & SL                      & 6.00\%           & 0.000                      & CL                      \\
          & $P_{in}$  & 99.41\%          & 0.000                      & SL                      & 0.00\%           & 0.000                      & CL                      \\
Gas Well  & $P_{ex}$ & 93.21\%          & 0.000                      & SL                      & 28.33\%          & 0.015                    & CL                      \\
          & $P_{in}$  & 51.33\%          & 0.344                      & SL                      & 44.67\%          & 0.521                      & SL                      \\
Well Log  & $P_{ex}$ & 95.09\%          & 0.000                      & SL                      & 7.22\%           & 0.000                      & CL                      \\
          & $P_{in}$  & 89.33\%          & 0.000                     & SL                      & 8.78\%           & 0.000                      & CL                      \\ \hline
\end{tabular}}
\end{center}
\end{table*}

\begin{figure*}
\centering
\includegraphics[width=0.9\textwidth]{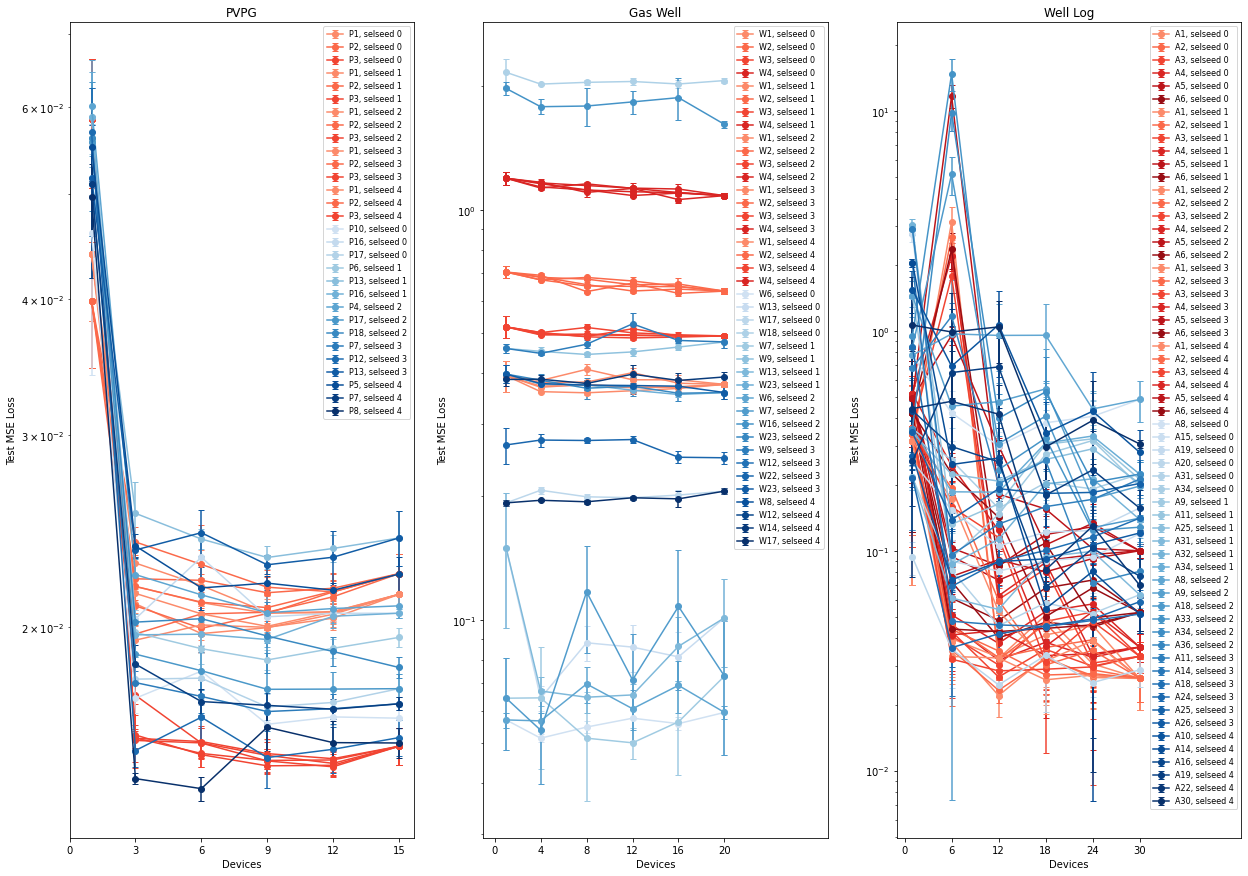}
\caption{Performance with various devices}\label{nodes}
\end{figure*}

\begin{figure*}
\centering
\includegraphics[width=0.9\textwidth]{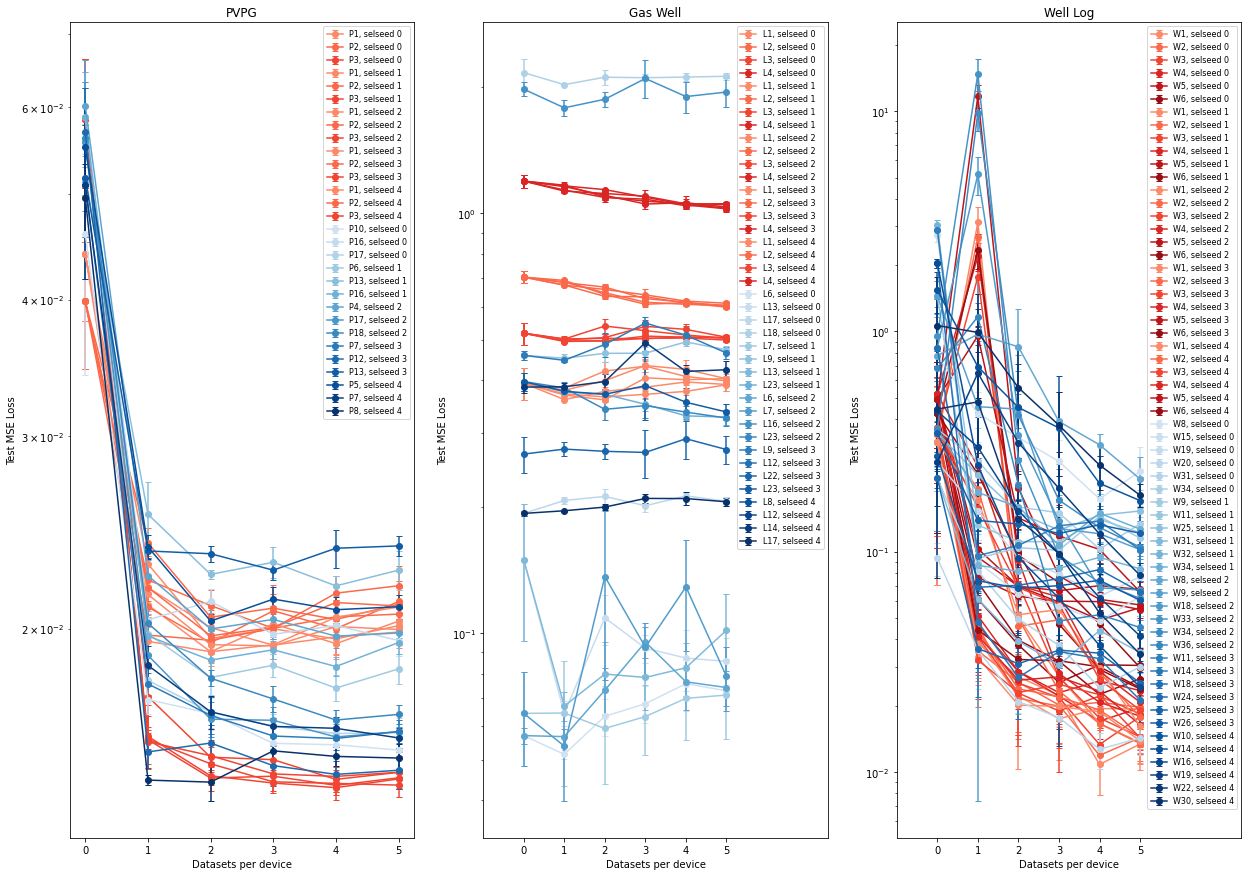}
\caption{Performance with various datasets}\label{sets}
\end{figure*}

The results of variations in data volume are shown in Figures \ref{nodes}, \ref{sets}, representing various participating devices scenario and owned datasets scenario respectively. The mean and standard deviation of test MSE are utilized to obtain the figure. Where "1" and "0" on the x-axis denotes local learning, respectively. Blue lines means performance in inside dataset, while red means outside dataset. The experiment result table can be found in Supplementary information. 

\subsubsection{PVPG Forecasting}
In the primary PVPG prediction experiment, SL outperforms Local Learning (LL) while falling short of Central Learning (CL) in both best account and mean value refer to Table \ref{result}, The distribution of SL test MSE versus LL and CL is obvious in Figure \ref{vs}, it can be obtained intuitively the same conclusion that, SL demonstrates superior performance over LL with local data, exhibiting a lower result distribution, while it doesn't match the performance of CL, which boasts an upper result distribution. The Mann-Whitney U Test yields a significantly small value (e-4 significance), supporting H1 that SL performs better than LL but worse than CL. Though SL did not outperforms CL, compared to CL, SL is more private and more secure, and the loss of accuracy is acceptable.

Both Figure \ref{nodes} and Figure \ref{sets} show an initial improvement followed by steady performance that standard deviation of test MSE is decreased. However, the performance get worse when data volume increased further, there is a trend that the test MSE mean value decrease then increase with data volume increasing.



\subsubsection{Gas Well Production Prediction}

SL demonstrates superior performance in external dataset of gas well prediction, owing to sufficient data, while providing marginal improvements on internal datasets and better generalization to unseen wells in Figure \ref{vs}. The result shows that the account of SL better than LL is 93.21\% in $P_{ex}$ while 51.33\% in $P_{in}$. The Mann-Whitney U Test p-value larger than threshold, means we cannot reject the hypothesis that the distributions of SL and LL test MSE in $P_{in}$ have no difference, although mean value of SL is better.

The example figure \ref{w1} shows that sufficient generalization can be obtained using LL, but there is an improvement in unseen wells in Figure \ref{nodes} and Figure \ref{sets}. While the improvement in gas well prediction isn't significant when data volume increasing, it notably reduces fluctuations. In this scenario, initially better-performing wells may deteriorate over time, while initially poorer-performing wells may improve.

Moreover, the magnitude of change on the internal test set isn't as significant as on the external dataset. There's a trend where performance on the internal test set decreases with an increase in the number of data groups, while performance on the external test set improves. This suggests that as heterogeneity between data groups increases, external generalization capability gradually improves, while internal generalization capability remains unchanged or decreases.

\subsubsection{Geophysical Well Log Generation}

SL outperforms LL in Figure \ref{vs}, from Table \ref{result}, we can obtain SL's mean test MSE value is better than LL and 95.09\% and 89.33\% results better in $P_{ex}$ and $P_{in}$. However, when data volume increasing is limited, heterogeneity between data due to the data selection method may lead to decreased learning effectiveness and increased Mean Squared Error (MSE), observed 7 out of 30 groups in Figure \ref{nodes} and Figure \ref{sets}. However, with an increase in the number of data groups and participating devices, the situation significantly improves.


\subsubsection{Local epoch}

\begin{figure*}
\centering
\includegraphics[width=0.8\textwidth]{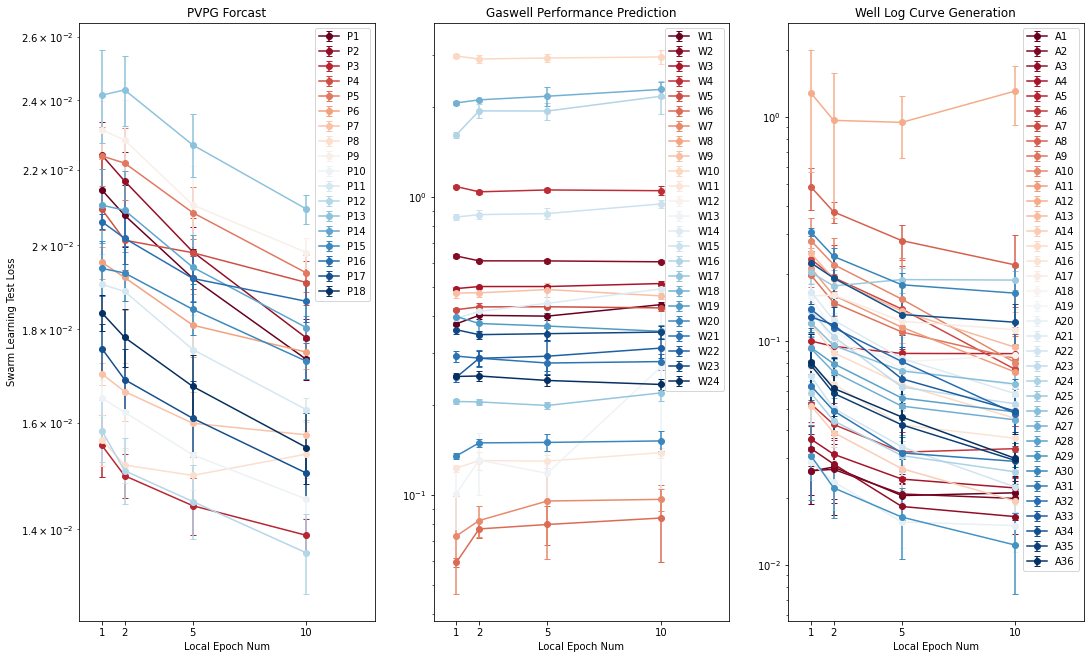}
\caption{Performance with various local epochs}\label{locep}
\end{figure*}

Additionally, experiment was designed to investigate the effect of local epoch size on model training. The local epoch size was incrementally increased from 1 to 2, 5, and 10, while the total number of epochs is kept constant. 5 random experiments conducted using data splits from the $1^{st}$ fold. The results showed in Figure \ref{locep} indicate that increasing the local epoch can further enhance the model's performance in Collaborated Learning scenarios where there is still room for improvement in model generalization performance, such as PVPG Forecast and Well Log Generation.

\subsection{Time, Security and Privacy Analysis}
\subsubsection{Time complexity Analysis}
The total time $t$ for Swarm Learning comprises three components: initialization time $t_{ini}$, time for local parameter updates $t_{locupdate}$, and time for model aggregation $t_{agg}$. Let's assume there are $E$ global epochs, and $k$ represents the number of devices participating in round $e$. The time for local updates is determined by the device with the longest duration in each round. This time can be further divided into parameter download time $t^{e,k}{pardownload}$, local training time $t^{e,k}{loctraining}$, and parameter upload time $t^{e,k}_{parupload}$. It's important to note that parameter uploading, downloading, and aggregation require consensus to be achieved on the blockchain.

\begin{equation}
t=t_{ini}+t_{locupdate}+t_{agg}
\end{equation}

\begin{equation}
  \begin{split}
  t_{locupdate}&=\sum^E_{e=0}max(t^{e,k}_{pardownload}\\
  &+t^{e,k}_{loctraining}+t^{e,k}_{parupload}) 
  \end{split}
\end{equation}

Table \ref{timetable} presents the outcomes of 100 epochs for 5 random cases. In scenarios where the training time for SL and FL are roughly equivalent, the communication cost between SL and FL varies by an order of magnitude. This discrepancy reflects the expenses associated with substituting central nodes, verifying node identities, achieving consensus among distributed nodes, and ensuring network structure security.

\begin{table}[!ht]
\caption{SL vs FL Time}\label{timetable}
\begin{center}
\resizebox{1\columnwidth}{!}{
\begin{tabular}{lllllll}
\hline
                                               & \multicolumn{2}{l}{PV} & \multicolumn{2}{l}{Gas Well} & \multicolumn{2}{l}{Logging}         \\ \cline{2-7} 
                                               & SL         & FL        & SL            & FL          & SL                         & FL     \\ \hline
$t_{locupdate}+t_{agg}(s)$                     & 3440.22    & 362.7     & 4198.87       & 287.29      & 4959.45& 207.93 \\
$max(\sum_{e=0}^E   t^{e,k}_{loctraining})(s)$ & 112.65     & 116.41    & 118.76        & 115.25      & 23.29   & 21.53  \\ \hline
\end{tabular}
}
\end{center}
\end{table}

\subsubsection{Security and Privacy Discussion}

In this section, we delve into the security and privacy considerations of SL in comparison to FL. Security pertains to the effective execution of the process, while privacy concerns whether user data is adequately safeguarded. 

We define "honest" participants as those who uploading correct model parameters and correctly executing the model parameter aggregation, relative to "malicious". "Curious" refers to participants attempting to obtain or infer user data. Federated Learning (FL), Federated Learning with single-key encryption (EFL)\cite{cheon2017homomorphic}, Federated Learning with multi-key encryption (Mk-EFL)\cite{mkckks_ma2022privacy}, and proposed Swarm Learning (SL) are compared. 


In the single-key homomorphic encryption FL process, all participants use the same pair of public and private key for parameters encryption and decryption. Each participant uses the public key to encrypt local data, generating ciphertext. The central server performs addition on the ciphertext from all participants, generating the aggregated ciphertext. The central server or participant uses the private key to decrypt the aggregated ciphertext to obtain the aggregated parameters. Steps are repeated for several iterations until the model converges or achieves satisfactory performance.

The multi-key homomorphic encryption FL process involves multiple participants, each participant using different pair of public and private key for encrypting and decrypting data to ensure data privacy and security. The weight parameters are encoded into plaintext and encrypted using local public keys to generate ciphertexts. Subsequently, these ciphertexts are uploaded to the parameter server. The central server aggregates the encrypted ciphertexts from all participants. Each participant then decrypts the aggregated encrypted ciphertext using their local private key and uploads the decrypted plaintext. The parameter server returns the aggregated plaintext, which is then decoded by the participants into the aggregated weight parameters.

Scenarios are designed to disscuss the security and privacy of four methods, the table is presented in Table \ref{ScrPrcy}, where "Y" indicates that the approach can manage the scenario, while "N" indicates cannot.

A malicious parameter server may disrupt the training process by halting services, incorrectly aggregating parameters, or leaking parameters externally. Additionally, a curious parameter server might analyze the data information embedded within the parameters. Federated learning methods reliant on servers, such as FL, EFL, and Mk-EFL, cannot address these scenarios. Even homomorphic encryption-based solutions are susceptible to parameter aggregation manipulation by adjusting the share of parameters from different participants. In data-sensitive energy service scenarios, the inference risks posed by parameter leakage, even when model parameters are encrypted, are unacceptable. Collective learning schemes overcome these issues by eliminating the need for external service providers. Parameter aggregation is executed and audited via code-based smart contracts, ensuring that erroneous operations cannot pass verification. The learning process remains uninterrupted despite some blockchain nodes ceasing to operate. By excluding external service providers, the risk of parameter leakage and analysis is effectively mitigated.

Malicious users may attempt to disrupt the federated learning training process by uploading incorrect parameters. Encrypted federated learning (EFL) and multi-key EFL (Mk-EFL) make it difficult to assess whether encrypted parameters are erroneous. Conversely, federated learning (FL) and secure learning (SL) can evaluate the correctness of encrypted parameters by analyzing their similarity. Curious users may seek to analyze parameters uploaded by other participants. In FL and EFL, collusion with the server grants access to other users' data. SL can be designed to mitigate this by preserving only metadata records while concealing parameters.

Regarding the security of blockchain networks, if parameters are not synchronized in a timely manner and forks exist, honest nodes of other organizations will not agree with the new block to ensure gradient descent, and collaborative learning records will not be overturned. If the packaged node is not honest, the consensus algorithm will reselect the packaged node. When a participant is malicious, that is, they want to disrupt the learning process, as long as most nodes are honest, it will not affect the learning process. However, in Federated Learning, the server has too much power. If it wants to disrupt the process through malicious actions, it will be relatively easy.

\begin{table*}[]
\tiny
\caption{Different scenario}\label{ScrPrcy}
\begin{center}
\resizebox{2\columnwidth}{!}{
\begin{tabular}{lllllll}
\hline
       & \multicolumn{2}{l}{Malicious   server} &                   & Curious server & Malicious client & Curious client \\ \cline{2-7} 
       & breakdown          & incorrectly integrating parameters         & parameter leakage & parameter inference          & parameter upload & parameter inference          \\ \hline
FL     & N             & N                      & N                 & N              & Y                & N              \\
EFL    & N             & N                      & N                 & N              & N                & N              \\
Mk-EFL & N             & N                      & N                 & Y              & N                & Y              \\
SL     & Y             & Y                      & Y                 & Y              & Y                & Y              \\ \hline
\end{tabular}
}
\end{center}
\end{table*}


\section{Conclusion}
In this paper, we design and implement a Swarm Learning scheme for series modeling in the energy sector, with the aim of obtaining high-performance machine learning models. The proposed Swarm Learning (SL) scheme addressed the security and privacy issues inherent in FL's centralized architecture. In this scheme, smart contracts and consensus mechanisms maintain consistent distribution during learning, ensuring parameters remain transparent, trustworthy, and immutable on-chain.

The effectiveness of our scheme is proved in three application scenarios, and the average performance of Swarm Learning is better than that of Local Learning in terms of cross-validation and repeated randomized trial performance. For gas well performance prediction scenarios, SL has a certain improvement due to the high test accuracy of the locally learned model, but the magnitude is not large. When nodes increases by the fixed number of data set and the number of sets increases by the fixed number of nodes, the overall performance is improved, but sometimes the effect deteriorates as the data volume increases. In addition, the influence of local epoch on the convergence performance of the model is also explored, the experimental results show that: improving the local epoch is a good means to improve the performance of the scenario when there is potential generalization.

Compared with FL, the training time of the proposed SL framework is about the same, but the cost difference of communication is about an order of magnitude, which is the price of substituting the central node, verifying the identity of the node, reaching consensus among distributed nodes, and ensuring the security of the network structure. Discussion indicates that SL provides better secure and privacy-preserving collaborative learning than FL, EFL and Mk-EFL.

For scenarios with large model parameters, the cost of storing the parameters on the blockchain and the network latency are high, so the off-chain storage scheme should be adopted, and we will focus on this issue in future research.











\section{Data and availability}
The data for this study is available from the corresponding author upon request.

\section{Declaration of Competing Interest}
The authors declare that they have no known competing financial interests or personal relationships that could have appeared to influence the work reported in this paper.

\section{Acknowledgments}
Jiang Chunbi was acknowledged for discussion and suggestion in ML. Deng Ruizhe was acknowledged to provide method of preprocessing PV data. Mu Ke, Wang Ziwei was acknowledged for discussion and suggestion in Blockchain.

\section{Supplementary information}

Supplementary information to this article, including dataset description, model details used in this paper, result tables, time analysis, security and privacy discussion, can be found in the online vision:

\appendix
\section{Dataset Description}\label{A}
\renewcommand\thetable{A.\arabic{table}}
\setcounter{table}{0}
The data descriptions of each dataset are shown in the Table \ref{D1},\ref{D3},\ref{D4}. For privacy reasons, the data is numbered anonymously.

\begin{table}[]
\scriptsize
\caption{Description of photovoltaic plants power generation load data}\label{D1}
\begin{center}
\resizebox{.9\columnwidth}{!}{
\begin{tabular}{lllllll}
\hline
    & count & mean       & std        & min       & 50\%       & max        \\ \hline
P1  & 6468  & 1065.97 & 1070.05 & 0       & 625.25   & 3817.00       \\
P2  & 6480  & 1200.97 & 1170.76 & 0       & 731.17   & 4203.00       \\
P3  & 6468  & 1347.59 & 1381.27 & 0        & 756.86   & 5673.50     \\
P4  & 6468  & 2087.36 & 2030.18 & 0       & 1270.88  & 7277.75    \\
P5  & 6468  & 1753.56 & 1656.44 & 0       & 1149.79  & 6130.00       \\
P6  & 6480  & 2402.42 & 2316.47 & 0       & 1585.69  & 8303.25    \\
P7  & 6480  & 2205.63 & 2154.89 & 0       & 1371.42  & 7934.20     \\
P8  & 6480  & 878.97  & 832.93  & 0       & 571.67   & 3206.10     \\
P9  & 6480  & 4225.77 & 4060.18 & 0       & 2636.25  & 14462.50    \\
P10 & 6480  & 4434.80 & 4149.57 & 0       & 2912.92  & 16354.55 \\
P11 & 6480  & 4057.71 & 3826.76 & 0       & 2578.75  & 14140.83 \\
P12 & 6468  & 988.39  & 932.54  & 0       & 637.33   & 3423.58 \\
P13 & 6468  & 1344.53 & 1297.90 & 0       & 847.55   & 4744.75    \\
P14 & 6480  & 2230.66 & 2089.42 & 0       & 1460.02  & 8057.90     \\
P15 & 6468  & 1860.88 & 1726.82 & 0        & 1237.11  & 7006.54 \\
P16 & 6480  & 2544.77 & 2453.21 & 0       & 1632.63  & 8938.13   \\
P17 & 6468  & 1750.38 & 1643.22 & 0       & 1076.32  & 5948.77 \\
P18 & 6468  & 1859.04 & 1702.35 & 0       & 1277.71  & 6168.91 \\ \hline
\end{tabular}}
\end{center}
\end{table}

\begin{table}[]
\scriptsize
\caption{Description of gas well production data}\label{D3}
\begin{center}
\resizebox{.8\columnwidth}{!}{
\begin{tabular}{lllllll}
\hline
          & count & mean      & std       & min & 50\%     & max      \\ \hline
W1  & 3744  & 19.10  & 9.18  & 0.00   & 20.24 & 61.84  \\
W2  & 3735  & 37.84 & 18.48  & 0.00   & 43.42 & 77.54  \\
W3  & 3236  & 42.06 & 15.37  & 0.00   & 47.25 & 70.03  \\
W4  & 3745  & 60.56  & 25.00 & 0.00   & 62.30 & 110.10 \\
W5  & 3745  & 53.12  & 30.13 & 0.00   & 56.14 & 102.56 \\
W6  & 3481  & 17.97 & 18.65  & 0.00   & 14.64 & 65.50   \\
W7  & 3590  & 23.52 & 17.45  & 0.00   & 20.31 & 73.73  \\
W8  & 2700  & 8.83  & 12.77  & 0.00   & 0.00  & 47.68  \\
W9  & 3715  & 23.44 & 16.19  & 0.00   & 20.63 & 71.95  \\
W10 & 3720  & 22.22 & 13.12  & 0.00   & 21.84 & 65.66  \\
W11 & 3720  & 22.38 & 22.34  & 0.00   & 19.21 & 81.77  \\
W12 & 3582  & 16.18 & 23.99  & 0.00   & 0.00  & 80.75  \\
W13 & 3713  & 14.68 & 17.56  & 0.00   & 9.14  & 60.64  \\
W14 & 3718  & 18.32 & 12.96  & 0.00   & 20.18 & 50.10     \\
W15 & 3725  & 21.58 & 16.40  & 0.00   & 20.82 & 59.74  \\
W16 & 440   & 15.53 & 2.66   & 0.00   & 15.80 & 25.18    \\
W17 & 3679  & 36.54 & 15.31  & 0.00   & 39.03 & 84.93  \\
W18 & 3661  & 88.09 & 33.23  & 0.00   & 98.63 & 136.63 \\
W19 & 3685  & 48.37 & 17.81  & 0.00   & 53.08 & 99.19   \\
W20 & 3715  & 50.01 & 22.49  & 0.00   & 50.03 & 97.03  \\
W21 & 3719  & 62.95 & 26.56  & 0.00   & 63.72 & 113.51 \\
W22 & 2875  & 15.94 & 12.28  & 0.00   & 19.20 & 60.85  \\
W23 & 3553  & 65.02 & 26.96  & 0.00   & 64.00 & 112.11 \\
W24 & 3554  & 73.75  & 34.58 & 0.00   & 72.61 & 133.93  \\ \hline
\end{tabular}}
\end{center}
\end{table}

\begin{table}[]
\scriptsize
\caption{Description of DTSM well log data}\label{D4}
\begin{center}
\resizebox{.8\columnwidth}{!}{
\begin{tabular}{lllllll}
\hline
    & count & mean   & std   & min     & 50\%     & max      \\ \hline
A1  & 2613  & 104.50 & 16.70 & 93.98   & 99.95    & 200.12   \\
A2  & 3731  & 104.34 & 16.17 & 87.63   & 99.85    & 206.70   \\
A3  & 2280  & 105.24 & 12.65 & 92.73   & 102.36   & 193.64   \\
A4  & 5546  & 104.07 & 14.72 & 89.45   & 100.00   & 202.55   \\
A5  & 9186  & 120.00 & 32.65 & 88.71   & 107.03   & 273.93   \\
A6  & 2154  & 102.96 & 15.04 & 89.53   & 99.86    & 188.60   \\
A7  & 3886  & 103.39 & 15.76 & 88.29   & 97.21    & 210.41   \\
A8  & 3677  & 102.79 & 15.05 & 89.82   & 98.13    & 208.02   \\
A9  & 2029  & 102.22 & 13.78 & 89.45   & 97.45    & 176.00   \\
A10 & 1094  & 113.59 & 15.57 & 96.84   & 108.09   & 181.90  \\
A11 & 1901  & 108.15 & 14.19 & 94.15   & 103.86   & 206.26   \\
A12 & 1911  & 102.56 & 12.92 & 89.79   & 98.90    & 169.61   \\
A13 & 1247  & 109.92 & 13.08 & 93.39   & 105.68   & 184.15   \\
A14 & 2108  & 104.21 & 15.86 & 91.82   & 100.06   & 195.42   \\
A15 & 11287 & 121.30 & 30.14 & 75.93   & 111.71   & 252.34   \\
A16 & 3891  & 103.13 & 16.69 & 89.61   & 99.13    & 211.89  \\
A17 & 3729  & 103.27 & 10.60 & 90.28   & 100.95   & 190.29   \\
A18 & 4223  & 103.68 & 13.64 & 89.47   & 99.00    & 199.23     \\
A19 & 17159 & 176.99 & 80.65 & 89.06   & 140.45   & 498.96   \\
A20 & 4024  & 102.30 & 12.54 & 89.36   & 98.50    & 199.01   \\
A21 & 4485  & 103.10 & 11.84 & 91.32   & 100.21   & 178.65  \\
A22 & 3542  & 104.57 & 10.15 & 90.77   & 102.39   & 188.30   \\
A23 & 18132 & 174.18 & 77.79 & 89.18   & 136.88   & 416.25    \\
A24 & 13958 & 185.57 & 81.97 & 84.90   & 162.11   & 359.63   \\
A25 & 18533 & 171.47 & 79.63 & 85.04   & 131.50   & 406.23  \\
A26 & 13186 & 139.46 & 54.20 & 66.41   & 116.56   & 332.59   \\
A27 & 14639 & 147.80 & 56.27 & 78.12   & 119.69   & 321.81    \\
A28 & 2277  & 104.46 & 16.57 & 91.86   & 98.97    & 193.59   \\
A29 & 4222  & 102.51 & 11.88 & 88.73   & 99.22    & 183.59  \\
A30 & 16923 & 160.58 & 65.80 & 76.15   & 126.95   & 357.58    \\
A31 & 2378  & 106.43 & 17.16 & 91.04   & 99.58    & 186.85   \\
A32 & 3918  & 104.57 & 16.55 & 92.01   & 99.35    & 205.33 \\
A33 & 3879  & 100.67 & 12.52 & 80.56   & 96.29    & 185.24  \\
A34 & 4303  & 100.88 & 12.71 & 86.42   & 96.95    & 196.85  \\
A35 & 4085  & 103.73 & 16.29 & 91.13   & 98.91    & 202.29   \\
A36 & 15619 & 182.49 & 80.77 & 90.46   & 160.16   & 411.64   \\ \hline
\end{tabular}}
\end{center}
\end{table}

\section{Dataset select method in data volume discussion}\label{B}
\renewcommand\thetable{B.\arabic{table}}
\setcounter{table}{0}

Known that "selseed" to control the variation in dataset selection. Taking one selseed for gas well performance prediction task as an example, we generate divided sets including all datasets in $P_{in}$, then delete dataset in every set, drop the null set after deletion, refer to Table \ref{tab nf}. For each selseed, we conduct repeat experiments using 5 fixed seeds to assess the randomness of the model.

\begin{table*}[]\caption{Dataset selection example of selseed 0 in gas well performance}\label{tab nf}
\begin{center}
    \resizebox{2\columnwidth}{!}{
\begin{tabular}{lll}
\hline
\multicolumn{1}{l}{} & Fix the number of participating devices                                                                         & Fix the number of datasets                                                                                                                                            \\ \hline
$P_{ex}$           & {[}0{]},{[}1{]},{[}2{]},{[}3{]}                                               & {[}0{]},{[}1{]},{[}2{]},{[}3{]}                                                                                                                 \\
$P_{in}$\#1                  & {[}5,6,10,19,20{]},{[}4,8,9,12,18{]},{[}7,15,16,21,23{]},{[}11,13,14,17,22{]} & {[}4{]},{[}5{]},{[}6{]},$\dots$,{[}23{]}                                                                                                  \\
$P_{in}$\#2                  & {[}5,6,19,20{]},{[}4,8,9,12{]},{[}7,15,16,23{]},{[}11,13,17,22{]}             & {[}4{]},{[}5{]},{[}6{]},{[}7{]},{[}8{]},{[}9{]},{[}10{]},{[}11{]},{[}12{]},{[}13{]},{[}15{]},{[}16{]},{[}17{]},{[}21{]},{[}22{]},{[}23{]} \\
$P_{in}$\#3                  & {[}5,19,20{]},{[}8,9,12{]},{[}15,16,23{]},{[}11,13,17{]}                      & {[}5{]},{[}8{]},{[}9{]},{[}11{]},{[}12{]},{[}13{]},{[}15{]},{[}16{]},{[}17{]},{[}21{]},{[}22{]},{[}23{]}                                  \\
$P_{in}$\#4                  & {[}5,20{]},{[}9,12{]},{[}16,23{]},{[}11,17{]}                                 & {[}5{]},{[}9{]},{[}11{]},{[}12{]},{[}16{]},{[}17{]},{[}21{]},{[}23{]}                                                                     \\
$P_{in}$\#5                  & {[}5{]},{[}12{]},{[}16{]},{[}17{]}                                            & {[}5{]},{[}12{]},{[}16{]},{[}17{]}                                                                                                        \\
\hline
\end{tabular}}
\end{center}
\end{table*}

\section{Example of Test Result}\label{C}

\renewcommand\thefigure{C.\arabic{figure}}
\setcounter{figure}{0}

\begin{figure*}
\centering
\includegraphics[width=1\textwidth]{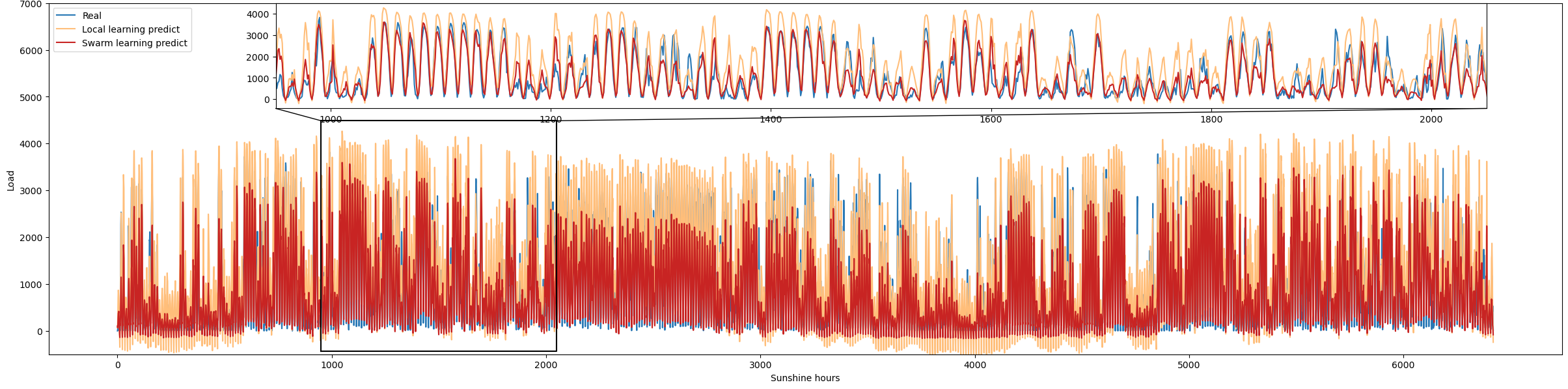}
\caption{P1 example result}\label{p1}
\end{figure*}

\begin{figure*}
\centering
\includegraphics[width=1\textwidth]{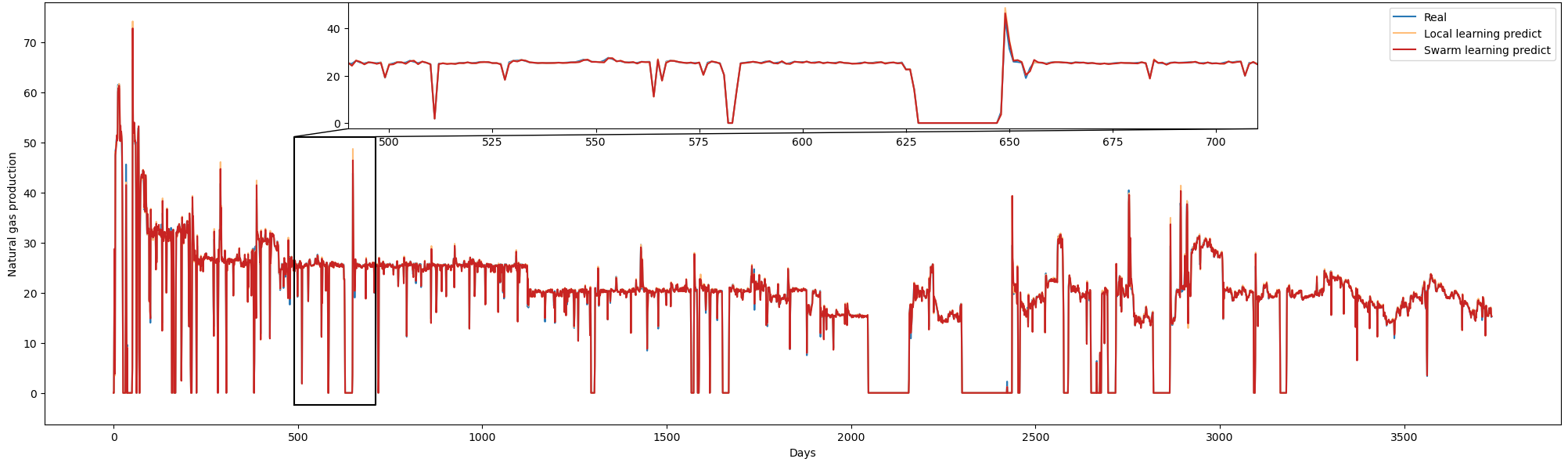}
\caption{W1 example result}\label{w1}
\end{figure*}

\begin{figure}
\centering
\includegraphics[width=0.9\columnwidth]{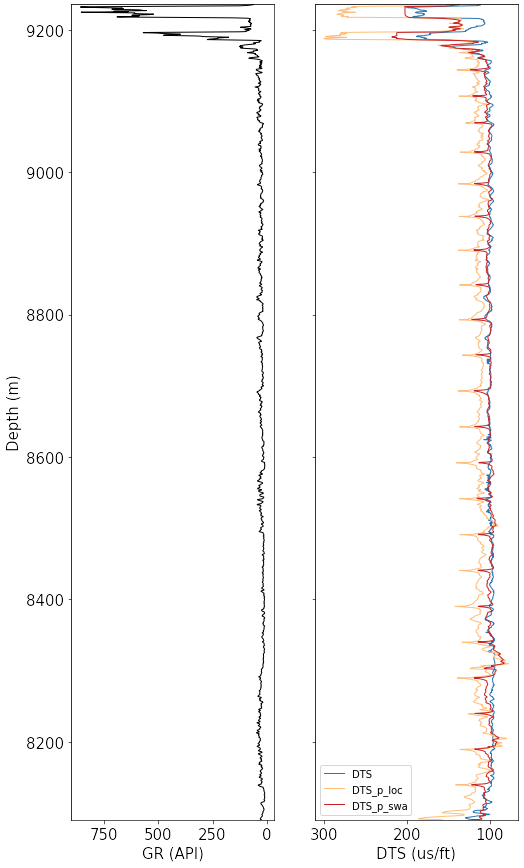}
\caption{A1 example result}\label{a1}
\end{figure}

The test results of external dataset P1, W1, A1, in the first fold and random seed 0 is shown here, as Figures \ref{p1}, \ref{w1}, and \ref{a1}.


\section{One-tailed Mann-Whitney U test}\label{D}
\setcounter{equation}{0}
\renewcommand\theequation{D.\arabic{equation}}
Here is a detailed explanation of the calculation method for conducting a one-tailed Mann-Whitney U test when the alternative hypothesis suggests that one group of data is smaller than the other. The null hypothesis (\(H_0\)) for the one-tailed Mann-Whitney U test states that there is no significant difference between the distributions of the two groups. The alternative hypothesis (\(H_1\)) suggests that one group of data is smaller than the other.

Calculation Steps:
1) Ranking: Combine the data from both groups and rank them from smallest to largest, assigning ranks to tied values by averaging the ranks.
2) Sum of Ranks: Calculate the sum of ranks (\(U\)) for each group.
3) Calculate the U Statistic: Determine the U statistic for the smaller group (\(U_s\)) using the formula:

\begin{equation}
    U_s = n_1 \cdot n_2 + \frac{n_1(n_1 + 1)}{2} - U_1
\end{equation}

where:
   \(n_1\) is the sample size of the smaller group.
   \(n_2\) is the sample size of the larger group.
   \(U_1\) is the sum of ranks for the smaller group.
4) Calculate the Critical Value: Determine the critical value from the Mann-Whitney U critical values table for the chosen significance level (\(\alpha\)) and sample sizes.
5) Compare U Statistic and Critical Value: Compare the calculated U statistic (\(U_s\)) with the critical value. If \(U_s\) is less than or equal to the critical value, reject the null hypothesis and accept the alternative hypothesis. Otherwise, fail to reject the null hypothesis.








\bibliographystyle{cas-model2-names}


\bibliography{sl_ref}

\end{document}